\documentclass[sigconf,nonacm]{acmart}

\renewcommand\footnotetextcopyrightpermission[1]{}
\setcopyright{none}
\usepackage{amsmath}
\usepackage{booktabs}
\usepackage{tabularx}
\usepackage{array}
\usepackage{makecell}
\usepackage{microtype}

\newcolumntype{Y}{>{\centering\arraybackslash}X}

\acmConference[Converted Manuscript]{Converted Manuscript}{2023}{Beijing, China}
\acmYear{2023}
\acmISBN{}
\acmDOI{}

\title{Relation Extraction Model Based on Semantic Enhancement Mechanism}

\author{Peiyu Liu}
\affiliation{%
  \institution{School of Computer Science (National Pilot Software Engineering School), Beijing University of Posts and Telecommunications}
  \city{Beijing}
  \postcode{100876}
  \country{China}
}

\author{Junping Du}
\authornote{Corresponding author.}
\email{junpingdu@126.com}
\affiliation{%
  \institution{Beijing Key Laboratory of Intelligent Telecommunication Software and Multimedia, Beijing University of Posts and Telecommunications}
  \city{Beijing}
  \postcode{100876}
  \country{China}
}

\author{Yingxia Shao}
\affiliation{%
  \institution{Beijing Key Laboratory of Intelligent Telecommunication Software and Multimedia, Beijing University of Posts and Telecommunications}
  \city{Beijing}
  \postcode{100876}
  \country{China}
}

\author{Zeli Guan}
\affiliation{%
  \institution{Beijing Key Laboratory of Intelligent Telecommunication Software and Multimedia, Beijing University of Posts and Telecommunications}
  \city{Beijing}
  \postcode{100876}
  \country{China}
}

\begin{document}

\begin{abstract}
Relational extraction is one of the basic tasks related to information extraction in the field of natural language processing, and is an important link and core task in the fields of information extraction, natural language understanding, and information retrieval. None of the existing relation extraction methods can effectively solve the problem of triple overlap. The CasAug model proposed in this paper based on the CasRel framework combined with the semantic enhancement mechanism can solve this problem to a certain extent. The CasAug model enhances the semantics of the identified possible subjects by adding a semantic enhancement mechanism. First, based on the semantic coding of possible subjects, pre-classify the possible subjects, and then combine the subject lexicon to calculate the semantic similarity to obtain the similar vocabulary of possible subjects. According to the similar vocabulary obtained, each word in different relations is calculated through the attention mechanism. For the contribution of the possible subject, finally combine the relationship pre-classification results to weight the enhanced semantics of each relationship to find the enhanced semantics of the possible subject, and send the enhanced semantics combined with the possible subject to the object and relationship extraction module. Complete the final relation triplet extraction. The experimental results show that, compared with the baseline model, the CasAug model proposed in this paper has improved the effect of relation extraction, and CasAug's ability to deal with overlapping problems and extract multiple relations is also better than the baseline model, indicating that the semantic enhancement mechanism proposed in this paper can further reduce the judgment of redundant relations and alleviate the problem of triple overlap.
\end{abstract}


\thanks{This work was supported by the Program of the National Natural Science Foundation of China (62192784, U22B2038, 62172056).}

\maketitle

\section{Introduction}

Relation extraction is one of the basic tasks related to information extraction in the field of natural language processing, which is mainly to identify the relationship between entities in text, and is one of the essential steps in downstream natural language processing tasks such as text comprehension and question answering. Relationship extraction is the extraction of entity relationship triplets from unstructured text to represent the relationships between two entities.

With the development and progress of technology, there are three main extraction methods in the field of relationship extraction: relationship extraction based on supervised learning, relationship extraction based on semi-supervised learning, and relationship extraction based on distant supervised learning. Due to the fact that the dataset used for the latter two relationship extraction methods is a collection of weakly labeled samples, the dataset contains a large amount of noise. Therefore, how to effectively denoise the dataset is one of the difficulties that these two relationship extraction methods need to overcome. For example, the multi-instance multi-label learning method proposed by Surdeanu et al. \cite{surdeanu2012multiinstance}, the attention mechanism proposed by Lin et al. \cite{lin2016neural}, and the DSGAN model constructed by Qin et al. \cite{qin2018dsgan} in combination with adversarial learning and sentence-level denoising are all designed to reduce the impact of error labels on performance in semi-supervised learning and remote supervised learning. Related semi-supervised short text classification work also uses heterogeneous graph attention to introduce typed semantic context into sparse text representations \cite{hu2019hgat}. Domain-specific artificial intelligence applications, including pediatrics, also require reliable extraction and organization of textual knowledge \cite{li2020pediatricAI}. In order to achieve a better extraction effect, this paper uses supervised learning to extract relations and makes full use of the labeled sample data in the public dataset to avoid the problem of data noise.

In relation extraction methods based on supervised learning, such as the pipeline extraction model proposed by Chan et al. \cite{chan2011syntactico}, there is a problem of error propagation; errors in the early entity recognition phase cannot be corrected in the later relationship classification phase. To address this issue, scholars have proposed extraction methods based on joint learning, such as the feature-based model proposed by Ren et al. \cite{ren2017cotype} and the neural network-based model proposed by Fu et al. \cite{fu2019graphrel}. Retrieval-oriented pre-training further shows that masked autoencoding can strengthen language representations for matching-heavy tasks \cite{xiao2022retromae}, and unified generative recommendation also shows that retrieval and ranking signals can be modeled in a single framework \cite{zhang2025generativeRecommendation}. However, these methods cannot effectively solve the problem of triple overlap, where different triples in a sentence have the same entity, and the same entity may appear in different relational triples. The problem of overlapping triples poses a huge challenge to relational classification methods.

In order to solve the problem of triple overlap and improve the accuracy of relationship extraction, this paper proposes the CasAug model. This model is based on the CasRel framework and introduces a semantic enhancement mechanism in the entity extraction stage. The extracted entities are semantically enhanced based on their relationship triplet types, and the semantically enhanced entities are fed into the relationship classification module to assist in the final relationship classification process. Since enhanced semantic representations may be used in downstream decision processes, interpretable machine learning research also suggests that learned features should remain understandable in intelligent decision scenarios \cite{li2019interpretable}; this is especially important for applied artificial intelligence scenarios such as pediatric decision support \cite{li2020pediatricAI}.

The main contributions of this paper include:
\begin{enumerate}
  \item In the subject extraction module, a semantic enhancement mechanism is introduced. This mechanism first pre-classifies the identified potential subjects, that is, pre-classifies their possible relationships. Then, based on the pre-classification results, the attention mechanism and subject vocabulary are combined to enhance the semantics of potential subjects.
  \item The loss function in the CasRel framework is improved by introducing the loss of relationship pre-classification in the semantic enhancement module, and a new label construction method is proposed for relationship pre-classification.
  \item The experimental results on public datasets show that the proposed CasAug framework has improved accuracy by 62.2\%, 26.1\%, 47.5\%, and 1.8\% compared to the baseline models NovelTagging, CopyRE, GraphRel, and CasRel, respectively.
\end{enumerate}

\section{Related Work}

There are two kinds of relation extraction algorithms based on supervised learning: pipeline learning \cite{zhou2005exploring,li2017tobit,cao2013review,meng2015formation,li2018resilient} and joint learning \cite{han2020more,kou2018hashtag,meng2013tracking,wei2019boosting,meng2015robust,li2017variance}. The pipeline-based method separates the two steps of relationship extraction, namely entity recognition and entity relationship recognition, and treats them as two independent tasks. Models are defined for each of them: first building a model for entity recognition, and then classifying relationships based on the entity recognition results \cite{socher2012semantic,wang2016relation,xu2015classifying,nguyen2015combining,li2016bilstm,cai2016bidirectional}. However, the pipeline-based approach ignores the correlation between two subtasks, resulting in the loss of some information. Additionally, errors in the entity recognition model will accumulate and propagate to the relational classification model, affecting the final extraction effect. Although cross-media retrieval is not relation extraction, semantics-adversarial and media-adversarial retrieval for scientific and technological information highlights the need to align heterogeneous semantic signals \cite{li2022crossmedia}; recent recommendation work similarly emphasizes the benefit of unifying retrieval and ranking objectives \cite{zhang2025generativeRecommendation}.

Compared with the pipeline-based relationship extraction method, the joint learning method can make full use of the interactive information between entities and relationships, complete the classification of entities to relationships in the process of entity extraction, and solve the problems of the pipeline method \cite{li2022stochastic,li2022vehiclefuel,li2023latenttopic,xiao2022lecf,li2022distributed,miwa2016end,katiyar2017going,chen2023speech,huang2021hgamn,shao2021memory}. Federated supervised cross-modal retrieval studies a related representation-alignment setting in which supervised semantic matching is learned across modalities without centralizing all data \cite{li2024fedCrossModal}. In the early days, Miwa et al. \cite{miwa2016end} completed joint representation of entities and relationships based on neural networks, and the LSTM unit sequence representation in the encoding layer was shared by entity recognition tasks and relationship classification tasks. However, the two subtasks in this model only share the bidirectional sequence LSTM representation of the encoding layer, and do not truly achieve joint learning between the two subtasks. Katiyar et al. \cite{katiyar2017going} improved Miwa's model by combining an attention mechanism and bidirectional LSTM, truly achieving joint extraction of entities and classification relationships. It is based on the shared encoding layer representation and the entity sequence representation output in the entity recognition subtask, and combines an attention mechanism to classify entity relationships.

Outside relation extraction, several representation-learning studies provide complementary evidence for the importance of multi-view and graph-aware semantics. Multi-view scholar clustering tracks dynamic interests across views \cite{li2023scholar}, while federated graph neural networks address representation learning for cross-graph node classification \cite{guan2021federated}. FedSIN further applies federated self-adaptive learning to information-network representation, indicating that network semantics can be adapted under distributed data constraints \cite{li2026fedsin}.

The above joint learning model based on parameter sharing has improved the problem of ignoring the relationship dependency between two subtasks and the existence of error accumulation and propagation in the pipeline method to a certain extent. Graph representation studies outside relation extraction likewise show why structural and semantic signals should be considered jointly: modularity-based deep community detection exploits graph community signals \cite{yang2016modularity}, T2-GNN tackles incomplete graph features and structures \cite{huo2023t2gnn}, and self-supervised graph co-training learns from complementary graph views \cite{xia2021graphCoTraining}. However, the above model discusses the two subtasks separately in the implementation process, and it still needs to conduct named-entity recognition first and then match the entities in pairs according to the entity recognition results. This pairwise matching method may generate redundant information between entities without any relationship. Zheng et al. \cite{zheng2017joint} proposed a new labeling strategy for relationship extraction, which introduces a unified labeling scheme to achieve joint decoding and transforms the task of extracting relationship triplets into an end-to-end sequence labeling problem. Due to the integration of entity and relationship information into a unified annotation scheme in this model, this method can directly model relationship triplet extraction at three levels. Wei et al. \cite{wei2020novel} proposed the CasRel model to solve the problem of triplet overlap, which constructs a new cascaded binary tagging framework. This framework no longer treats relationships as discrete labels like the previous model, but instead models relationships as functions that map subjects in sentences to objects, thereby solving the problem of triplet overlap. This framework proposes the concept of cascading, which means that after identifying the subject, it will fall back to the semantic encoding layer, read the semantic encoding of the identified subject, and send it to the object and relationship recognition module. Lightweight sequential recommendation with filter-enhanced MLPs further illustrates how feature filtering can reduce redundant sequential signals \cite{zhou2022filterMLP}, while generative recommendation provides another example of reducing pipeline separation between retrieval and ranking \cite{zhang2025generativeRecommendation}.

However, there are still many redundant relationships that need to be determined in this model. Relying solely on the output of the subject recognition module and the semantic encoding of possible subjects based on cascading extraction as inputs to the object and relationship recognition module to extract relationship triplets may not be sufficient, nor can it effectively narrow the range of possible relationships, nor is it sufficient to classify relationships in a more detailed manner. Semantic augmentation for named entity recognition also shows that additional semantic signals can improve extraction in noisy text \cite{nie2020semantic}. Scientific-publication representation learning with semantic-similarity attention and hypergraph convolution further supports the use of similarity-aware attention to capture higher-order semantic relations \cite{li2026semanticHypergraph}. To sum up, this paper adds a semantic enhancement module based on the CasRel framework, builds the CasAug model, and readjusts the loss function. This model first pre-classifies the relationships of the identified possible subjects based on the hierarchical system, and then combines the pre-classification results and the subject dictionary library to enhance the semantics of the identified possible subjects, further enriching the input of the object and classification module, assisting in the relationship extraction process, and reducing the judgment of redundant relationships.

\section{Model}

\subsection{CasRel Framework}

\begin{figure}
  \centering
  \includegraphics[width=\columnwidth]{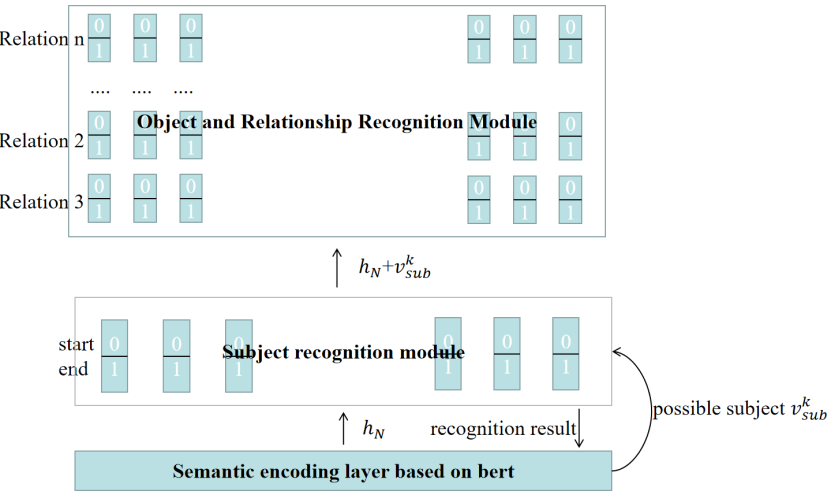}
  \caption{CasRel Framework Diagram}
  \Description{Framework diagram of the CasRel model.}
  \label{fig:casrel}
\end{figure}

CasRel is a relationship extraction framework proposed by Wei et al. \cite{wei2020novel}, which is generally divided into two modules: subject recognition module and object-relationship recognition module. The basic idea is to extract relationship triplets through cascading steps. Firstly, possible subjects are extracted from the input sentence in the subject recognition module, and then the semantic encoding of these possible entities is found in the semantic encoding layer. Combined with the output of the subject recognition module and the extracted semantic encoding of possible subjects, they are sent to the object and relationship recognition module. In this module, all possible relationships are checked for each candidate subject, and the module determines whether the relationship can combine the subject with the object in the sentence. If so, the extraction of the relationship triplet is successful. Figure~\ref{fig:casrel} shows the overall framework of CasRel.

As shown in Figure~\ref{fig:casrel}, the bottom layer of the CasRel framework is the semantic encoding layer, which uses a BERT encoder to extract feature information $X_j$ from sentence $X_j$ and input it into the subsequent subject recognition module. The BERT encoder \cite{devlin2019bert} is a language representation model based on multi-layer bidirectional transformers, which learns the deep feature representation of words by jointly adjusting the left and right contexts of each word. The subject recognition module recognizes all possible subjects in a sentence by directly decoding the encoding vector $h_N$ generated by the BERT encoder. This module uses binary 0/1 tagging to detect the start and end positions of the subject, assigning two binary tags to each word to indicate whether the word corresponds to the start or end position of a particular subject. As shown in Equations~\eqref{eq:start-s} and \eqref{eq:end-s},
\begin{align}
p_i^{start\_s} &= \sigma(W_{start}x_i+b_{start}), \label{eq:start-s}\\
p_i^{end\_s} &= \sigma(W_{end}x_i+b_{end}). \label{eq:end-s}
\end{align}
Among them, $p_i^{start\_s}$ and $p_i^{end\_s}$ respectively represent the probability of recognizing the $i$-th word in the input sentence as the starting and ending positions of a certain subject. If the probability exceeds a certain threshold, the corresponding binary marker bit is set to 1; otherwise it is set to 0. $x_i$ is the vector representation of the $i$-th word in the input statement encoding, that is, $x_i=h_N[i]$. $W(\cdot)$ is the trainable weight, $b(\cdot)$ represents the bias term, and $\sigma$ is the activation function.

After the subject recognition module identifies which words are the subject, it goes back to the semantic encoding layer to obtain the semantic encoding $v^k_{sub}$ of these words, and then inputs the output $x_i$ of the subject recognition module and the semantic encoding $v^k_{sub}$ of these words together into the object and relationship recognition module for the final extraction of relationship triplets.

The object and relationship recognition module identifies the object and its relationship from the sentence based on the extracted subject. This module sets an object recognizer for each relationship, as shown in Equations~\eqref{eq:start-o} and \eqref{eq:end-o}:
\begin{align}
p_i^{start\_o} &= \sigma(W^r_{start}(x_i+v^k_{sub})+b^r_{start}), \label{eq:start-o}\\
p_i^{end\_o} &= \sigma(W^r_{end}(x_i+v^k_{sub})+b^r_{end}). \label{eq:end-o}
\end{align}
Among them, $p_i^{start\_o}$ and $p_i^{end\_o}$ represent the probability of marking the $i$-th word in the input sequence as the starting and ending position of the object, and $v^k_{sub}$ represents the semantic encoding of the $k$-th subject detected in the subject recognition module. $W^r(\cdot)$ represents the trainable weight of relationship $r$, $b^r(\cdot)$ represents the bias term of relationship $r$, and $\sigma$ is the activation function. For each type of relationship, the object recognizer determines whether the relationship can recognize the corresponding object based on the extracted subject. If so, the relationship triplet extraction is successful.

Compared with other classic relationship extraction models, the CasRel framework can solve the problem of triple overlap to a certain extent and improve the accuracy of relationship classification. However, through analysis, it was found that there are still many redundant relationships to be determined in the object and relationship recognition module. Relying solely on the semantic encoding of possible subjects to identify objects and relationships in this module may not be sufficient and cannot effectively narrow the scope of possible relationships. Therefore, this article adds a semantic enhancement mechanism in the subject recognition module. After identifying potential subjects, a pre-classification process is performed. Based on the pre-classification results, semantic enhancement is combined with the subject dictionary library to enrich the input of the object and relationship module, reduce the judgment of redundant relationships, and further solve the problem of triple overlap.

\subsection{CasAug Framework Based on Semantic Enhancement Mechanism}

\begin{figure}
  \centering
  \includegraphics[width=\columnwidth]{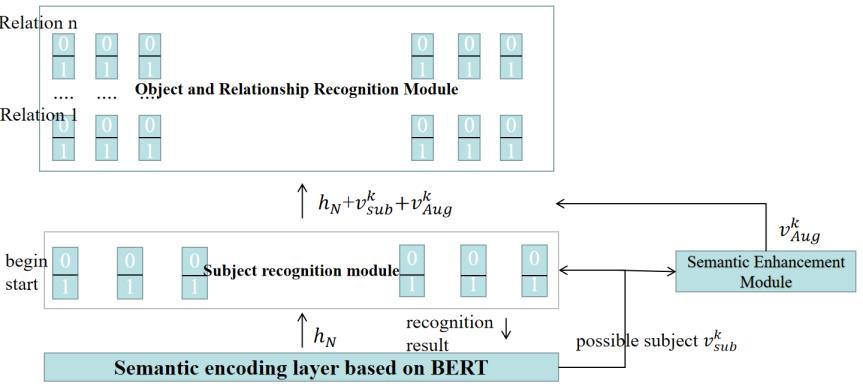}
  \caption{CasAug Framework Diagram}
  \Description{Framework diagram of the CasAug model.}
  \label{fig:casaug}
\end{figure}

\begin{figure}
  \centering
  \includegraphics[width=\columnwidth]{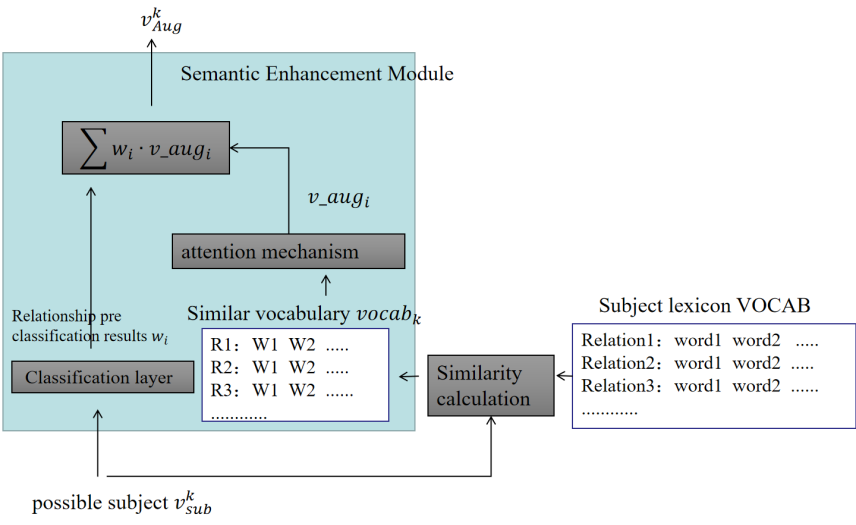}
  \caption{Semantic Enhancement Mechanism}
  \Description{Detailed diagram of the semantic enhancement mechanism.}
  \label{fig:semantic}
\end{figure}

Figure~\ref{fig:casaug} shows the CasAug framework diagram. Based on CasRel, the CasAug model adds a semantic enhancement module to the subject recognition module. Figure~\ref{fig:semantic} shows the detailed diagram of the semantic enhancement module proposed in this article. As shown in the detail diagram, the input of this module is $v^k_{sub}$, which is the semantic encoding of the $k$-th possible subject identified by the subject recognition module, and the output is $v^k_{aug}$, which is the enhanced semantic vector of the input vector $v^k_{sub}$. This module first performs relationship pre-classification on $v^k_{sub}$, and then calculates the semantic enhancement vector $v\_aug_{num\_rel_i}$ that contributes to the potential subject $v^k_{sub}$ under each relationship based on the subject lexicon and attention mechanism. Combining the pre-classification results $w_{num\_rel_i}^k$ and $v\_aug_{num\_rel_i}$, the final enhanced semantic information for the potential subject $v^k_{sub}$ can be obtained.

$v^k_{sub}$ flows to two branches in total. The first branch passes through a classification layer, through which a relationship pre-classification is performed on the possible subject $v^k_{sub}$, that is, based on the semantic encoding of $v$, trying to determine the probability $w_{num\_rel_i}$ of $v^k_{sub}$ belonging to each relationship category, where $0<i\le num\_rel$ and $num\_rel$ is the total number of relationships.

The second branch flows to the similarity calculation module, which calculates the first $n$ words that are similar to the possible subject $v^k_{sub}$ under different relationship categories based on the subject lexicon $VOCAB$, and obtains a similar vocabulary $vocab_k$ for the possible subject, where $vocab_k\in R^{num\_rel\times n\times 768}$.

\textbf{Subject lexicon $VOCAB$.} The subject lexicon collects word vector information of the subject words belonging to each relationship, which is calculated by BERT's pre-trained model. Before the model starts training, for each relationship in the dataset, we need to collect a certain number of subject words that can form a relationship triplet with that relationship. The BERT pre-training model calculates the word vectors of these words to form the subject lexicon $VOCAB$, and then calculates the similarity with the possible subject word $v^k_{sub}$.

\textbf{Semantic similarity calculation.} In this article, we use the cosine similarity method to calculate the semantic similarity between the possible subject $v^k_{sub}$ and the words in the subject vocabulary. The formula is as follows:
\begin{equation}
d^{m_i}_{num\_rel_j}=
\sqrt{\sum_{x=1}^{768}(v^k_{sub\_x}-v^{m_i}_{num\_rel_j\_x})^2}.
\label{eq:semantic-distance}
\end{equation}
Among them, $v^k_{sub\_x}$ represents the $x$-th position of the possible subject word vector $v^k_{sub}$, $v^{m_i}_{num\_rel_j\_x}$ represents the $x$-th position of the $m_i$-th word vector in the $num\_rel_j$-th relationship of the subject lexicon $VOCAB$, and $d^{m_i}_{num\_rel_j}$ represents the semantic similarity between the possible subject word $v^k_{sub}$ and the $m_i$-th subject word in the $num\_rel_j$-th relationship of the subject lexicon $VOCAB$, where $0<num\_rel_j\le num\_rel$ and $0<m_i\le m$.

\textbf{Similar vocabulary $vocab_k$.} After calculating the semantic similarity between the possible subject $v^k_{sub}$ and each subject in the subject lexicon, based on this calculation result, for each relationship in the lexicon, $n$ topic words with the closest possible subject semantics in each relationship are selected to construct a similar vocabulary $vocab_k$.

After obtaining a list of similar words for possible subject $v^k_{sub}$, not all of these calculated similar words contribute to semantic enhancement of word $v^k_{sub}$ in that context. Therefore, in this article, we use an attention mechanism \cite{lin2016neural} to distinguish the contribution of different words to semantic enhancement of possible subject $v^k_{sub}$ in a specified context.

Firstly, for the similar words under each relationship in the similar vocabulary $vocab_k$, it is necessary to calculate the weights of these similar words, that is, the different contributions of these similar words to the semantic enhancement of the possible subject word $v^k_{sub}$ in the specified context. The calculation formula is shown in Equation~\eqref{eq:attention-weight}:
\begin{equation}
p^{n_j}_{num\_rel_i}=
\frac{\exp(v^k_{sub}\cdot v^{n_j}_{num\_rel_i})}
{\sum_{j=1}^{n}\exp(v^k_{sub}\cdot v^{n_j}_{num\_rel_i})}.
\label{eq:attention-weight}
\end{equation}
Among them, $v^{n_j}_{num\_rel_i}$ is the $n_j$-th similar word of $v^k_{sub}$ in the $num\_rel_i$-th relationship in the similarity vocabulary $vocab_k$, and $p^{n_j}_{num\_rel_i}$ is the calculated contribution of $v^{n_j}_{num\_rel_i}$ to the semantic enhancement of the possible subject word $v^k_{sub}$.

After calculating the contribution degree of all similar words in a certain relationship, based on the contribution degree, the semantic enhancement vector of all words in that relationship to the possible subject $v^k_{sub}$ can be calculated, as shown in Equation~\eqref{eq:rel-aug}:
\begin{equation}
v\_aug_{num\_rel_i}=\sum_{n_j=1}^{n}p^{n_j}_{num\_rel_i}\cdot v^{n_j}_{num\_rel_i}.
\label{eq:rel-aug}
\end{equation}
Finally, by combining the relationship pre-classification results $w_{num\_rel_i}$ calculated from the first branch, the final semantic enhancement information $h^k_{Aug}$ for possible subject $v^k_{sub}$ can be calculated, as shown in Equation~\eqref{eq:haug}:
\begin{equation}
h^k_{Aug}=\sum_{num\_rel_i=0}^{num\_rel}w_{num\_rel_i}\cdot v\_aug_{num\_rel_i}.
\label{eq:haug}
\end{equation}
Finally, the output $h^k_{Aug}$ of the semantic enhancement module serves as the input of the object and relationship recognition module, and together with the possible subject $v^k_{sub}$, assists in the extraction of the final relationship triplet:
\begin{align}
p_i^{start\_o} &= \sigma(W^r_{start}(x_i+v^k_{sub}+h^k_{Aug})+b^r_{start}), \label{eq:aug-start}\\
p_i^{end\_o} &= \sigma(W^r_{end}(x_i+v^k_{sub}+h^k_{Aug})+b^r_{end}). \label{eq:aug-end}
\end{align}

\subsection{Loss Function of the Model}

Finally, the loss function $Loss_{CasRel}$ of the model is shown in Equation~\eqref{eq:loss}:
\begin{equation}
Loss_{CasRel}=loss(s|x_j)+loss(o|x_j)+loss(o_{\emptyset}|s,x_j)+loss(w_i|s).
\label{eq:loss}
\end{equation}
The definitions of $loss(s|x_j)$, $loss(o|x_j)$ and $loss(o_{\emptyset}|s,x_j)$ refer to the framework of CasRel \cite{wei2020novel}, which respectively represent the loss function of the subject recognition module and the object and relationship recognition module. $loss(w_i|s)$ represents the loss caused by pre-classification of possible subjects in the semantic enhancement mechanism, and the calculation process is shown in Equation~\eqref{eq:pre-loss}:
\begin{equation}
loss(w|s)=Mean(binary\_crossentropy(target_s-Dense(v^k_{sub}))).
\label{eq:pre-loss}
\end{equation}
Where $binary\_crossentropy$ represents the binary cross entropy function, $Mean$ represents the averaging function, and $Dense$ represents the classification layer, where the input is the possible subject $v^k_{sub}$ and the output is the pre-classification result of the possible subject $v^k_{sub}$. $target_s$ is the probability label of the relationship to which the possible subject $v^k_{sub}$ belongs.

Due to the problem of repeating triplets, each subject may belong to different relationships; that is, the same subject may appear in different relationship triplets. Therefore, when constructing the label $target_s$, we count $target\_count_s$, which is the number of times each subject in the dataset belongs to different relationships, as shown in Equation~\eqref{eq:target-count}:
\begin{equation}
target\_count_s=[c_i]\quad(0<i\le num\_rel).
\label{eq:target-count}
\end{equation}
Among them, $c_i$ represents the number of times the subject $s$ appears in the $i$-th relationship. By normalizing $target\_count_s$, the probability label of this subject is constructed:
\begin{equation}
target_s=\frac{1}{\sum c_i}target\_count_s=\left[\frac{1}{\sum c_i}c_i\right].
\label{eq:target}
\end{equation}

\section{Experimental Result}

\subsection{Experimental Setup}

\textbf{Dataset.} In this article, we use the public dataset NYT \cite{riedel2010modeling} to train the proposed model. The NYT dataset consists of 1.18 million sentences and 24 predefined relationships, and the sentences in this dataset typically contain multiple relationship triplets. Therefore, the NYT dataset is very suitable for evaluating the effectiveness of models in extracting overlapping relationship triplets.

In this article, we used 56,195 pieces of data from the NYT dataset for training and 5,000 pieces of data for testing. Based on the overlap patterns of different triplets, the sentences in the test set were divided into three categories: normal test set, entity-to-entity overlap test set (EPO), and single entity overlap test set (SEO). More information on the dataset can be found in Table~\ref{tab:nyt-data}.

\begin{table}
  \caption{NTT Data Information}
  \label{tab:nyt-data}
  \centering
  \begin{tabularx}{\columnwidth}{YYY}
    \toprule
    Category & Training set & Testing set \\
    \midrule
    Normal & 37013 & 3266 \\
    EPO & 9782 & 978 \\
    SEO & 14735 & 1297 \\
    ALL & 56195 & 5000 \\
    \bottomrule
  \end{tabularx}
\end{table}

\textbf{Evaluation indicator.} According to previous work \cite{zheng2017joint}, only when each part of the relationship triplet is extracted correctly can the extraction of the relationship triplet be considered successful. In this article, we compare the proposed model with the baseline model through precision (Prec.), recall (Rec.), and F1 value (F1).

\textbf{Comparison method.} In this article, we use the following four models as baselines to compare with the proposed framework: NovelTagging \cite{zheng2017joint}, CopyRE \cite{zeng2018extracting}, GraphRel \cite{fu2019graphrel}, and CasRel \cite{wei2020novel}.

NovelTagging proposes a new labeling strategy for entities and relationships, transforming the joint extraction task in relationship extraction into a sequence labeling problem for the first time. CopyRE is the first to integrate the decoder-encoder architecture into the relationship extraction task, solving the problem of triple overlap by generating all relationship triples in sentences. GraphRel extracts overlapping relationships through graph convolutional neural network, decomposes entity pairs into multiple word pairs, and considers the prediction of all word pairs. CasRel first identifies all possible subjects in a sentence by building a subject recognition module and an object and relationship recognition module, and then identifies all possible relationships and corresponding objects based on possible entities.

\subsection{Experimental Result}

The performance of the five methods on the global test set is shown in Table~\ref{tab:test-set}. It can be seen that among the five models, NovelTagging performs the worst, and the model proposed in this article, CasAug, performs the best in accuracy. Compared with the baseline model, the CasAug model has an accuracy improvement of 62.2\%, 26.1\%, 47.5\%, and 1.8\% on the test set compared to NovelTagging, CopyRE, GraphRel, and CasRel, respectively.

\begin{table}
  \caption{Test Set Results}
  \label{tab:test-set}
  \centering
  \begin{tabularx}{\columnwidth}{YYYY}
    \toprule
    Method & Prec & Rec & F1 \\
    \midrule
    NovelTagging & 0.562 & 0.626 & 0.593 \\
    CopyRE & 0.723 & 0.678 & 0.699 \\
    GraphRel & 0.621 & 0.594 & 0.607 \\
    CasRel & 0.896 & 0.893 & 0.897 \\
    \textbf{CasAug} & \textbf{0.912} & \textbf{0.895} & \textbf{0.896} \\
    \bottomrule
  \end{tabularx}
\end{table}

In order to further validate the model's ability to handle overlapping problems, in this article, we divided the test set into three categories: ordinary test set, entity-to-entity overlapping test set (EPO), and single entity overlapping test set (SEO). The F1 values of the four models on the test set are shown in Table~\ref{tab:different-tests}.

\begin{table}
  \caption{Results from Different Test Sets (The Data for CopyRE and GraphRel in the Table Is Directly Referenced from CasRel)}
  \label{tab:different-tests}
  \centering
  \begin{tabularx}{\columnwidth}{YYYY}
    \toprule
    Method & Normal & SEO & EPO \\
    \midrule
    CopyRE & 0.660 & 0.486 & 0.550 \\
    GraphRel & 0.696 & 0.512 & 0.528 \\
    CasRel & 0.873 & 0.912 & 0.918 \\
    \textbf{CasAug} & \textbf{0.914} & \textbf{0.915} & \textbf{0.922} \\
    \bottomrule
  \end{tabularx}
\end{table}

From Table~\ref{tab:different-tests}, it can be seen that the CasAug model proposed in this article performs best on all three test sets. Among them, CasAug performs best on the EPO test set, with improvements of 67.6\%, 75.3\%, and 0.43\% compared to CopyRE, GraphRel, and CasRel, respectively. This indicates that the model proposed in this article can solve the problem of entity overlap to a certain extent.

In order to further analyze the model's ability to extract multiple relationships, based on the number of relationship triplets contained in a sentence, we divided the test set into different types, where $N$ refers to the number of relationship triplets contained in a sentence.

\begin{table}
  \caption{Test Results When $N$ Is Different (The Data for CopyRE and GraphRel in the Table Is Directly Referenced from CasRel)}
  \label{tab:n-different}
  \centering
  \begin{tabularx}{\columnwidth}{YYYYY}
    \toprule
    $N$ & 1 & 2 & 3 & 4 \\
    \midrule
    CopyRE & 0.671 & 0.586 & 0.520 & 0.536 \\
    GraphRel & 0.710 & 0.615 & 0.574 & 0.551 \\
    CasRel & 0.852 & 0.901 & 0.921 & 0.932 \\
    \textbf{CasAug} & \textbf{0.832} & \textbf{0.910} & \textbf{0.930} & \textbf{0.944} \\
    \bottomrule
  \end{tabularx}
\end{table}

From Table~\ref{tab:n-different}, it can be seen that compared with other baseline models, the CasAug framework proposed in this paper has the best performance, and as $N$ increases, the F1 value of CasAug becomes higher, indicating that the effectiveness of relationship extraction in this model will not decrease due to the complexity of relationships in sentences.

\section{Conclusion}

This article proposes a relationship extraction model CasAug based on semantic enhancement mechanism in conjunction with the CasRel framework. This model extracts possible subjects from the subject recognition module and sends them to the semantic enhancement mechanism module to extract the enhanced semantics of the possible subjects. In the semantic enhancement module, firstly relationship pre-classification is performed on possible subjects based on their semantic encoding. Then, semantic similarity calculation is performed on the subject lexicon to obtain a similar word list for possible subjects. Based on the obtained similar word list, the contribution of each word in different relationships to possible subjects is calculated through an attention mechanism. Finally, based on the pre-classification results of relationships, the enhanced semantics of each relationship are weighted to obtain the final enhanced semantics of the possible subject. The enhanced semantics are combined with the possible subject and sent to the object and relationship extraction module to complete the final relationship triplet extraction. In addition, CasAug adjusted the original loss function, introduced the loss of relationship pre-classification in the semantic enhancement module, and proposed a new tag construction method for relationship pre-classification. The experimental results show that compared with the baseline model, the CasAug model proposed in this paper has improved performance in relation extraction, and CasAug's ability to handle overlapping problems and extract multiple relationships is also better than the baseline model. This indicates that the semantic enhancement mechanism proposed in this paper can further reduce the judgment of redundant relationships and solve the problem of triple overlap.

\bibliographystyle{gbt7714-numerical}
\bibliography{references}

\end{document}